\documentclass{article}

\usepackage{PRIMEarxiv}

\usepackage[utf8]{inputenc} % allow utf-8 input
\usepackage[T1]{fontenc}    % use 8-bit T1 fonts
\usepackage{hyperref}       % hyperlinks
\usepackage{url}            % simple URL typesetting
\usepackage{booktabs}       % professional-quality tables
\usepackage{amsfonts}       % blackboard math symbols
\usepackage{nicefrac}       % compact symbols for 1/2, etc.
\usepackage{microtype}      % microtypography
\usepackage{lipsum}
\usepackage{fancyhdr}       % header
\usepackage{graphicx}       % graphics
\graphicspath{{media/}}     % organize your images and other figures under media/ folder
\usepackage{amsmath}
\title{Geospatial Artificial Intelligence for Satellite-Based Flood Extent Mapping: Concepts, Advances, and Future Perspectives}

\author{
  Hyunho Lee, Wenwen Li \\
  School of Geographical Sciences and Urban Planning \\
  Arizona State University \\
  Tempe, AZ, USA \\
  \texttt{\{hlee401, wenwen\}@asu.edu} \\
}

\begin{document}
\maketitle

\section{Synonyms}
GeoAI for Satellite-Based Flood Detection, GeoAI for Satellite-Based Flood Segmentation

\section{Definition}

Geospatial Artificial Intelligence (GeoAI) \cite{li2024geoai, li2020geoai} for satellite-based flood extent mapping systematically integrates artificial intelligence techniques with satellite data to identify flood events and assess their impacts, for disaster management and spatial decision-making \cite{bentivoglio2022deep, cova1999gis}. The primary output often includes flood extent maps, which delineate the affected areas, along with additional analytical outputs such as uncertainty estimation and change detection.

\section{Historical Background}

Geographic Information Science (GIScience) provides the theoretical foundations for spatial analysis methods and modeling based on various types of geospatial data such as satellite imagery, unmanned aerial vehicle (UAV) imagery, digital elevation models (DEM), and Light Detection and Ranging (LiDAR) data. These spatial analysis methods and models have been widely applied in emergency management, including disaster monitoring and prediction, hazard zone assessment, resource optimization, and damage evaluation. Among various natural disasters, floods are one of the most socioeconomically significant events \cite{tellman2021satellite}. To support effective flood response, flood extent mapping methods have been developed within GIScience for disaster management by integrating diverse geospatial data to provide timely and accurate information for all phases of flood management: mitigation, preparedness, response, and recovery \cite{ajmar2017response}.

Traditionally, rule-based approaches utilizing remote sensing indices, such as the Normalized Difference Water Index (NDWI) \cite{mcfeeters1996use} and the Modified Normalized Difference Water Index (MNDWI) \cite{xu2006modification}, have been applied to flood extent mapping. However, advancements in satellite technology, including the increase in the number of satellites for Earth observation, higher-resolution imagery, and shorter revisit cycles, have significantly increased the volume of available satellite data. This expansion has led to the development of new methods that can effectively extract information and recognize patterns from large-scale satellite data \cite{li2020geoai}.

Artificial Intelligence (AI) is a broad field dedicated to developing computational systems that emulate human intelligence and problem-solving capabilities. Machine Learning (ML), a subset of AI, focuses on identifying patterns in large-scale data and using these patterns to make predictions. As a subfield of ML, Artificial Neural Networks (ANNs) are computational models inspired by the structure and mechanisms of biological neural networks in animal brains \cite{brahme2014comprehensive}. With the introduction and refinement of the backpropagation algorithm, ANNs have evolved into a powerful tool for various applications by effectively modeling complex and nonlinear relationships between input and output data through end-to-end optimization. 

In the 2010s, Convolutional Neural Networks (CNNs), which are classifiers using an ANN model combined with feature extractors designed specifically for image analysis, achieved state-of-the-art performance in image recognition tasks, outperforming traditional computer vision algorithms \cite{li2022geoai}. One of the contributing factors that made this breakthrough possible is the Rectified Linear Unit (ReLU) \cite{nair2010rectified} activation function, which enabled the efficient training of deep neural networks with many layers, a key characteristic of Deep Learning (DL). Afterward, the ResNet architecture \cite{he2016deep} further enhanced the efficiency of training deeper neural networks by introducing skip connections, which directly connect the activations of one layer to subsequent layers, bypassing intermediate layers. Even after that, various deep learning model architectures, such as Generative Adversarial Networks (GANs) \cite{goodfellow2014generative}, Graph Convolutional Networks (GCNs) \cite{kipf2016semi}, Transformer \cite{vaswani2017attention}, and Vision Transformer (ViT) \cite{dosovitskiy2020image}, have been introduced. These advances in deep learning models have led to the in-depth application of GeoAI in the geospatial domain, particularly in flood extent mapping \cite{lee2024improving, li2023assessment}.

\section{Scientific Fundamentals}

The study of GeoAI for satellite-based flood extent mapping bridges the knowledge of flood-affected area characteristics in satellite data with an understanding of how AI, particularly deep learning models, leverages these features. In this section, we will outline the characteristics of Synthetic Aperture Radar (SAR) and Multispectral Imaging (MSI) data, the primary data sources used in satellite-based flood extent mapping. Then, we will briefly explain the optimization principles of deep learning by illustrating the learning process of ANNs and CNNs. Finally, advancements in GeoAI for this research field will be discussed.

\subsection{Data sources}

SAR data are the primary source for satellite-based flood extent mapping, offering observations of the Earth's surface in all weather conditions, regardless of day or night \cite{uddin2019operational, ajmar2017response, boccardo2014remote, chaouch2012synergetic}. This functionality stems from SAR sensors' ability to detect scattered energy from emitted microwave pulses. The intensity of scattered energy is primarily determined by surface roughness characteristics \cite{grimaldi2020flood}. While rough terrain surfaces generate high backscatter by dispersing energy in multiple directions including back to the sensor, open water surfaces produce low backscatter by reflecting radar signals away from the sensor. However, relying solely on SAR data for flood extent mapping faces some limitations, including speckle noise, difficulty distinguishing man-made flat surfaces (e.g., tarmac) from open water, and double-bounce backscattering from buildings and flooded vegetation \cite{amitrano2024flood, grimaldi2020flood}. On the other hand, MSI data, although limited by cloud cover in terms of observational availability, provides valuable water-sensitive spectral bands such as Near Infrared (NIR) and Shortwave Infrared (SWIR) during clear-sky conditions which significantly enhance the accuracy of flood extent mapping \cite{konapala2021exploring}. Furthermore, due to ease of visual interpretation, MSI data are predominantly utilized to assess flood-induced damage to infrastructure, such as buildings and roads \cite{ajmar2017response, boccardo2014remote}.

\subsection{AI methods}

\begin{figure}[t]
  \centering
  \includegraphics[width=1.0 \linewidth]{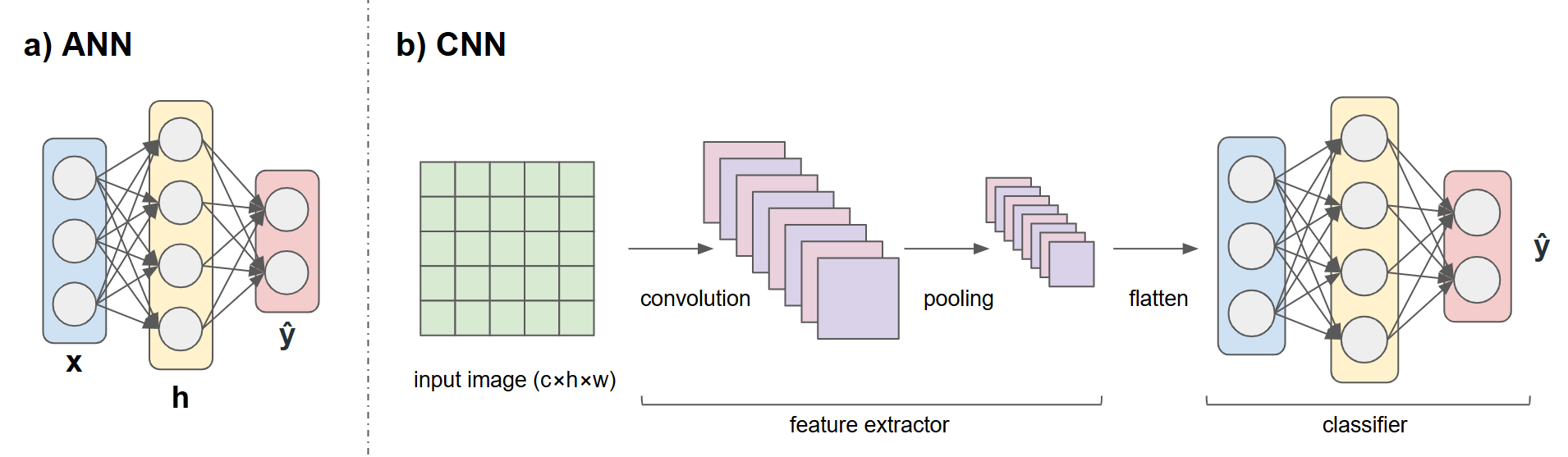}
  \caption{Architectures of AI models. (a) ANNs, (b) CNNs.}
  \label{fig:fig1}
\end{figure}

\begin{figure}[t]
  \centering
  \includegraphics[width=0.7 \linewidth]{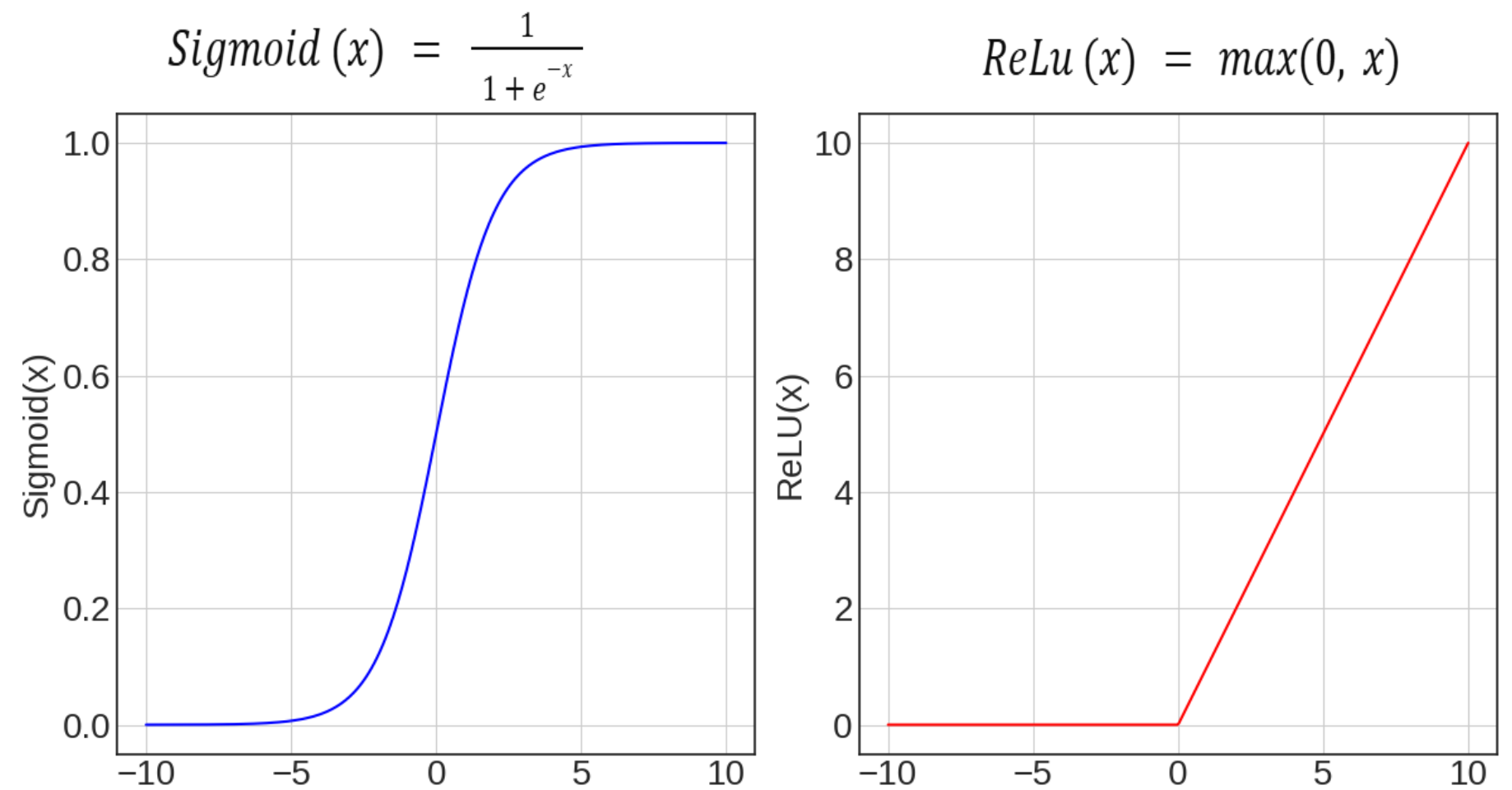}
  \caption{Activation functions in ANNs.}
  \label{fig:fig2}
\end{figure}

ANNs are AI models composed of multiple perceptrons or neurons in each layer as shown in Fig. \ref{fig:fig1} (a). In ANNs, the model's output (or prediction) is calculated through a feedforward process. Specifically, in the feedforward process, identical operations are performed at each neuron as follows. Let the input value to the neuron be $x_1$, $x_2$, $...$, $x_n$, the weights corresponding to these inputs be $w_1$, $w_2$, $...$, $w_n$, and the bias term be $b$. The neuron calculates the weighted sum of the inputs, adds the bias, and then applies an activation function $\sigma$. $z$ is the input to the activation function, which is the weighted sum of inputs plus the bias, and h is the output of the neuron after the activation function is applied (see Eq. (\ref{eq:eq1}) and Eq. (\ref{eq:eq2})). The activation function $\sigma$ can be a variety of functions, such as Sigmoid and ReLU (see Fig. \ref{fig:fig2}). 

\begin{align}
    z = \sum_{i=1}^n w_i x_i + b
    \label{eq:eq1} 
\end{align}

\begin{align}
    h = \sigma(z)
    \label{eq:eq2} 
\end{align}

The ANNs compute the output $\widehat{Y}$ through the feedforward process. Then, the difference, denoted as $L$, between the predicted output $\widehat{Y}$ and the actual ground truth $Y$ is calculated using a loss function (see Eq. (\ref{eq:eq3})). Specifically, for classification tasks, cross-entropy is the primarily utilized loss function, and mean squared error is most common for regression (see Eq. (\ref{eq:eq4}) and Eq. (\ref{eq:eq5})).

\begin{align}
   L = \textit{loss\_function} (\widehat{Y}, Y)
   \label{eq:eq3} 
\end{align}

\begin{align}
    \textit{CrossEntropy} (\widehat{Y}, Y) = \sum_{i=1}^C y_i \log(\hat{y_i})
    \label{eq:eq4} 
\end{align}

\begin{align}
    \textit{MeanSquaredError} (\widehat{Y}, Y) = \frac{1}{N}\sum_{i=1}^N (y_i - \hat{y_i})^2
    \label{eq:eq5} 
\end{align}

The calculated loss value is used to adjust weights during the backpropagation process. The backpropagation process is carried out using the chain rule to obtain the gradients ($\frac{\partial L}{\partial w_i}$) of the loss value $L$ with respect to the weights. This step determines which direction each weight should be adjusted (positive or negative) to minimize the loss value, with the magnitude of the adjustment controlled by the learning rate ($\eta$). Subsequently, based on the calculated gradients, the weights are updated through gradient descent algorithm (see Eq. (\ref{eq:eq6})). In training ANNs, learning rate is a hyperparameter that influences the model's training process. Additionally, various optimizers, which control weight updates, have been studied. One of the frequently utilized optimizers is Adam \cite{kingma2014adam}, which improves upon the gradient descent algorithm.

\begin{align}
    w_i^{new} = w_i^{old} - \eta \frac{\partial L}{\partial w_i}
    \label{eq:eq6} 
\end{align}

CNNs are an extended architecture of ANNs that combine a feature extractor and a classifier to process unstructured data, typically images, as presented in Fig. \ref{fig:fig1} (b). Particularly, in CNNs, ANNs are utilized as the classifier, and convolutional layers and pooling layers are leveraged for feature extraction. The convolutional layer performs a convolution operation between an input feature map and a filter (also known as a kernel). The role of the convolutional layer is to extract local patterns (e.g., edges, textures) from the input image. The pooling layer is used to downsample the spatial dimensions (height and width) of the feature map. The most common type of pooling is max pooling, where the maximum value in each local region is selected. The convolutional layer includes trainable weights for each filter, whereas the pooling layer generally does not include any weights. The learning process of a CNN also includes iterative steps of feed forward and backpropagation to optimize the model's weights by minimizing the difference between model’s predictions and labels. In GeoAI for satellite-based flood extent mapping, deep learning models for segmentation tasks, such as CNN-based models (e.g., U-Net \cite{ronneberger2015u}, DeepLabV3+ \cite{chen2018encoder}) and Vision Transformers (e.g., SegFormer \cite{xie2021segformer}), are primarily employed and optimized in an end-to-end manner, following the same principles as described in this section.

Vision Transformers (ViTs) \cite{dosovitskiy2020image} have emerged as an alternative architecture to CNNs for image processing tasks, including segmentation. In comparison to Convolutional Neural Networks (CNNs), which are designed to extract local features via convolutional layers, Vision Transformers (ViTs) are characterized by their ability to capture global dependencies through the self-attention mechanism. In segmentation, ViTs divide the input image into fixed-size patches, which are then flattened and projected into an embedding space. These patch embeddings are combined with positional encodings to retain spatial information and fed into a Transformer encoder. The encoder consists of multiple layers of multi-head self-attention and feed-forward networks, enabling the model to learn both local and global contextual relationships across the image. For segmentation tasks, ViT-based architectures like SegFormer \cite{xie2021segformer} integrate hierarchical feature representations and lightweight decoders to produce high-resolution segmentation maps.

Masked Autoencoders (MAE) \cite{he2022masked} introduced a self-supervised learning approach for Vision Transformers (ViTs) by randomly masking a portion of image patches and training the model to reconstruct the missing parts. This method allows the model to learn robust feature representations without labeled data. During training, the model infers missing patches from the visible ones, fostering the learning of generalizable features by leveraging contextual information. The MAE framework comprises an encoder that processes visible patches and a decoder that reconstructs the missing patches to approximate the original image. After pretraining, the encoder can be fine-tuned for downstream tasks, such as segmentation, by leveraging the feature representations learned during self-supervised training. This approach is particularly advantageous for satellite-based flood extent mapping, where labeled data may be scarce, as it allows the model to learn generalized patterns from large amounts of unlabeled satellite imagery \cite{li2023assessment}.

\begin{figure}[t]
  \centering
  \includegraphics[width=1.0 \linewidth]{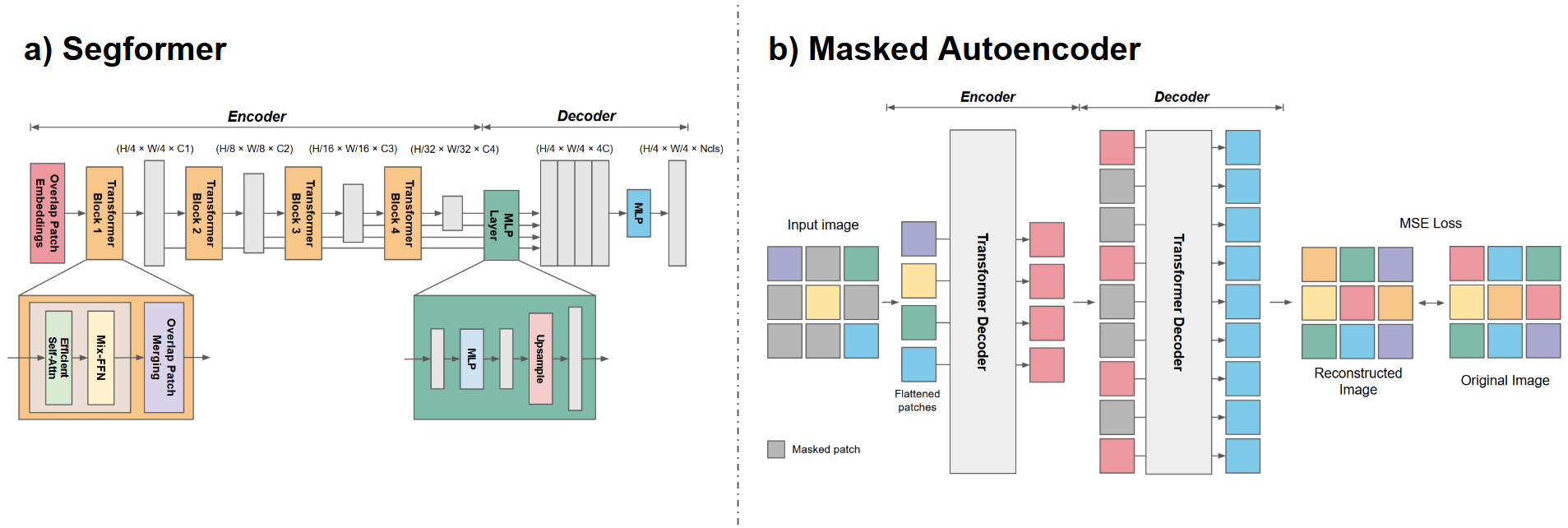}
  \caption{Architectures of ViT-based models. (a) Segformer \cite{xie2021segformer}; (b) Masked Autoencoder (MAE) \cite{he2022masked}.}
  \label{fig:fig3}
\end{figure}

\subsection{GeoAI for satellite-based flood extent mapping}

\begin{figure}[t]
  \centering
  \includegraphics[width=0.8 \linewidth]{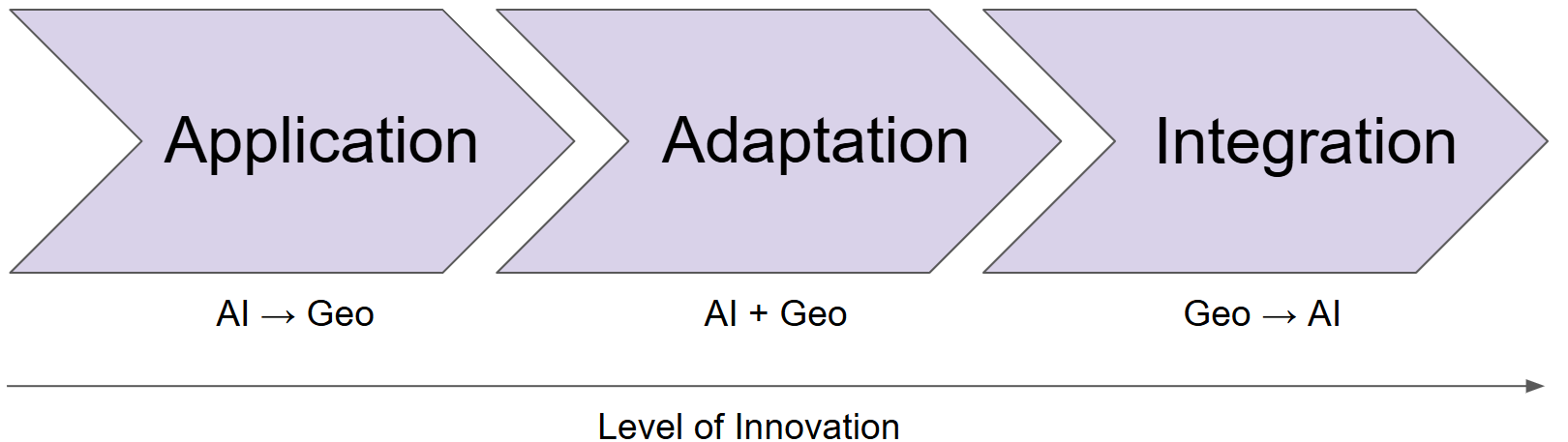}
  \caption{Level of innovation in GeoAI \cite{li2022geoai}.}
  \label{fig:fig4}
\end{figure}

The innovation of GeoAI for satellite-based flood extent mapping has followed a trajectory similar to that of GeoAI research in general, evolving through phases of application, adaptation, and integration, as illustrated in Fig. \ref{fig:fig4}. In 2018, CNNs were first applied to satellite data for flood extent mapping \cite{nogueira2018exploiting, kang2018flood} due to their similarity in recognizing patterns within images. Studies applying CNNs to flood extent mapping demonstrated their superior performance compared to conventional methods, including rule-based and machine learning approaches \cite{nogueira2018exploiting, kang2018flood, wieland2019modular, gebrehiwot2019deep, nemni2020fully, dong2021monitoring, katiyar2021near, bereczky2022sentinel, li2023u}. 

Around the same time, since 2018, researchers have published studies that include benchmark datasets for flood extent mapping \cite{khouakhi2022need}. This is closely related to the essential role of training data in the learning process of CNNs. Afterward, since 2020, benchmark datasets that incorporate multimodal geospatial data as input have been published \cite{bonafilia2020sen1floods11, rambour2020flood, Cloud_to_Street2022, montello2022mmflood, drakonakis2022ombrianet, zhang2023new, he2023cross}. Even after that, benchmarking datasets composed of time series SAR data, as well as datasets specifically designed for detecting flood extents in urban areas, are being released for flood extent mapping \cite{bountos2023kuro, zhao2024urbansarfloods}. These publicly available benchmark datasets have further accelerated GeoAI research for flood extent mapping. For instance, since the release of the Sen1Floods11 dataset \cite{bonafilia2020sen1floods11}, subsequent flood extent mapping studies have been conducted using this benchmark dataset with a focus on improving the accuracy and efficiency of flood extent mapping \cite{wang2024multi, lee2024improving, li2023assessment, garg2023cross, konapala2021exploring, katiyar2021near, bai2021enhancement}.

\section{Key Applications}

\subsection{Flood monitoring and early warning system}

\begin{figure}[t]
  \centering
  \includegraphics[width=0.8 \linewidth]{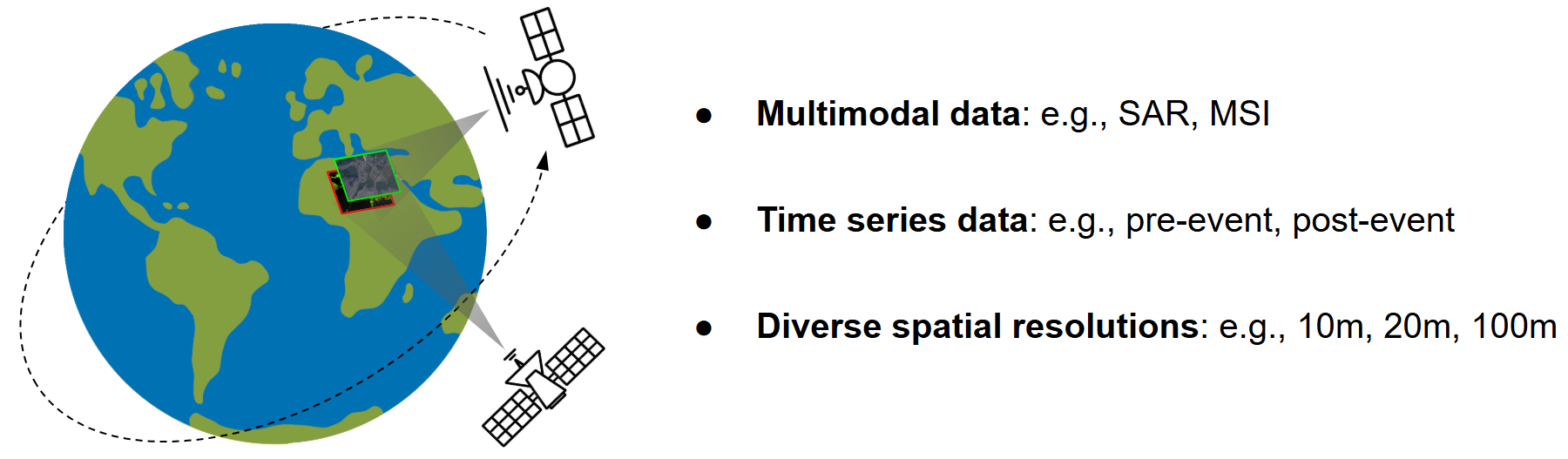}
  \caption{Unique characteristics of satellite data, adapted from \cite{rolf2024position} with modifications.}
  \label{fig:fig5}
\end{figure}

Flood monitoring and early warning systems are essential for effectively responding to the flood events. In this context, research has focused on two key applications: integrating the unique characteristics of geospatial data into deep learning models and addressing practical constraints in flood management operations. Recent advancements in GeoAI for satellite-based flood extent mapping have led to the development of deep learning models that explicitly integrate the unique characteristics of geospatial data, such as multimodality \cite{konapala2021exploring, munoz2021local, kim2021synergistic}, time-series \cite{peng2019patch}, and varying spatial resolutions \cite{zhang2021flood} (see Fig. \ref{fig:fig5}). Additionally, there are studies that improve mapping performance by utilizing both multimodality and time-series characteristics \cite{drakonakis2022ombrianet, he2023cross}. Furthermore, to reduce the burden of labeled data annotation for training deep learning models, research on weakly supervised learning \cite{bonafilia2020sen1floods11}, unsupervised learning \cite{akiva2021h2o, li2019urban}, active learning \cite{lee2024improving}, and transfer learning \cite{li2023assessment} has been explored in the context of GeoAI for flood extent mapping. In addition to CNNs, research is also focused on the application and enhancement of computer vision architectures, such as ViT, for flood extent mapping using satellite data \cite{saleh2024dam}.

On the other hand, for operational flood monitoring, the study in \cite{wieland2019modular} introduced a processing chain that covers the required modules for satellite-based flood extent mapping. This study demonstrated that deep learning models can be effectively integrated into the operational flood monitoring framework. However, in practical satellite-based flood extent mapping scenarios, the satellite data acquisition process accounts for the majority of the total process time. Therefore, it is critical to anticipate the demand for satellite images and request them in advance to provide timely flood extent maps \cite{wania2021increasing}. This is because satellite images have a large file size, while the data transmission bandwidth between the satellite and the ground station is limited. To address this challenge, the study in \cite{mateo2021towards} proposed a machine learning model for flood mapping with onboard processing capabilities, offering a solution to mitigate satellite data transmission constraints by transmitting processed flood maps instead of raw satellite data. As such, while flood monitoring and early warning systems are positioned as key applications of GeoAI for flood extent mapping, successful implementation in real-world flood management scenarios requires careful consideration of practical constraints to address operational challenges.

\subsection{Flood damage estimation}

 Flood damage estimation is a critical component of disaster management, enabling rapid damage assessment, prioritization of recovery efforts, and monitoring of progress. Previous studies have primarily utilized methods that generate flood extent maps and combine them with existing Land Use and Land Cover (LULC) maps to estimate flood damage \cite{arun2020flood, glas2016flood,wouters2021improving}. This approach includes overlaying flood extent maps, which delineate the spatial distribution of flooded areas, with LULC data to identify affected land use types, such as residential areas, agricultural fields, infrastructure, and transportation networks. While this method provides a broad overview of the impacted regions and is relatively straightforward to implement, it has some limitations. Since LULC data is often static, it may not reflect real-time changes or dynamic conditions during a flood event, such as temporary land use alterations. Additionally, differences in spatial resolution between the flood extent maps and LULC data can influence the accuracy of this approach, potentially leading to either an underestimation or an overestimation of flood damage.
 
Recently, there has been a shift toward leveraging deep learning models to detect flood damage at a more granular level, such as identifying damaged roads and buildings \cite{hansch2022spacenet, sakamoto2024proposal}. These models are trained using labeled pre- and post-flood satellite imagery, enabling them to directly compare and detect changes in infrastructures caused by flooding. This approach offers significant advantages, including the ability to identify subtle damage patterns, such as partially collapsed structures or eroded roads, which may be missed by traditional overlay methods. Furthermore, deep learning models can capture intricate damage patterns from large-scale satellite data, making them well-suited for assessing damage in compound disaster scenarios, such as hurricanes and floods. However, this method also faces challenges, such as the need for extensive labeled datasets for training, which can be time-consuming and resource-intensive to create.

\section{Future Directions}

GeoAI for flood extent mapping has shown significant improvements in extracting patterns from large volumes of satellite data and enhancing prediction accuracy compared to previous methods, especially from an automation perspective. However, there are still several open questions and challenges to be addressed. Therefore, in this section, we summarize potential research directions to improve the accuracy, reliability, and efficiency of GeoAI for flood extent mapping.

\subsection{Application-driven innovation in GeoAI for flood extent mapping}

While prior GeoAI studies on satellite-based flood extent mapping have primarily focused on improving deep learning models by incorporating the unique characteristics of geospatial data, their deployment in operational scenarios is equally, if not more, critical for addressing real-world challenges. Real-world flood management requires overcoming practical barriers such as handling missing or incomplete satellite data, generating additional flood-related information (e.g., inundation levels and flow velocity) from flood maps, accurately mapping urban floods in complex environments, and incorporating uncertainty estimation to account for data and model limitations. These challenges present opportunities for advancing GeoAI in flood extent mapping. Addressing deployment constraints through innovative research is essential for both academic progress and practical implementation \cite{rolnick2024application}.

\subsection{Integrating physical theories and domain knowledge into GeoAI for flood extent mapping}

Current GeoAI models primarily rely on data-driven learning methods, which may struggle to predict patterns that are not represented in training data \cite{bentivoglio2022deep}. A promising direction is the integration of physical theories and domain knowledge into GeoAI models to enhance their predictive capabilities \cite{li2024using, xu2023spatial}. This knowledge-driven approach ensures predictions are both physically consistent and domain-aware. Moreover, it is particularly valuable for improving model reliability in extreme climate scenarios, where data-driven models may fail to generalize effectively.

\subsection{Explainable GeoAI for flood extent mapping}

Deep learning models are generally designed as black-box systems, where the relationship between inputs and outputs is determined by automatically adjusting the model's weights, making their internal workings challenging to interpret \cite{hsu2023explainable}. To address this, ongoing research in the field of GeoAI is focused on developing methods to interpret the outcomes of these models \cite{li2024geoai, hsu2023explainable}. In image classification tasks, eXplainable AI (XAI) techniques such as Shapley Additive exPlanations (SHAP) \cite{lundberg2017unified}, Gradient-weighted Class Activation Mapping (Grad-CAM) \cite{selvaraju2017grad}, Layer-wise Relevance Propagation (LRP) \cite{bach2015pixel}, and Integrated Gradients (IG) \cite{sundararajan2017axiomatic} have been introduced to quantify and visualize the reasoning behind a model's predictions. Future research in flood mapping can benefit from advances in XAI, particularly in developing explainability techniques tailored for complex models and tasks, such as segmentation tasks and transformer-based models. 

In conclusion, GeoAI has unlocked new opportunities to enhance the accuracy and efficiency of satellite-based flood extent mapping. Moving forward, developing more explainable models, integrating domain knowledge more deeply, and driving application-focused innovations will be key to advancing GeoAI solutions for timely flood mapping in support of disaster management and decision-making.

%Bibliography
\bibliographystyle{unsrt}  
\bibliography{references}  

\end{document}